# Distributed Processing of Biosignal-Database for Emotion Recognition with Mahout

Varvara Kollia, Oguz H. Elibol

**Abstract**—This paper investigates the use of distributed processing on the problem of emotion recognition from physiological sensors using a popular machine learning library on distributed mode. Specifically, we run a random forests classifier on the biosignal-data, which have been pre-processed to form exclusive groups in an unsupervised fashion, on a Cloudera cluster using Mahout. The use of distributed processing significantly reduces the time required for the offline training of the classifier, enabling processing of large physiological datasets through many iterations.

―――――――――――――――― ◆ ――――――――――――――――

## 1 INTRODUCTION

There is a lot of interest today in emotion recognition, as it is an area with very broad applications to a large number of systems, such as human-computer interaction (HCI) systems, machine emotional intelligence, health industry etc. Emotion recognition is intrinsically a very challenging problem, with its main difficulties stemming from the fact that the problem itself is very subjective and hard to quantify. There are many popular emotion definitions and models, both in terms of discrete emotion subsets, as well as mappings in two- and three-dimensional spaces. In this work, we assume a 3D emotional model.

With increasing interest in the area, new and large datasets are being collected, enabling new insights to be discovered in the area. These datasets necessitate distributed processing for enhanced scalability and performance. Popular distributed machine learning libraries can augment the process of training accurate classifiers offline, to build prediction models based on large amounts of data. We used Mahout on distributed mode to train a random forest classifier, on the DEAP dataset. Using a distributed approach allowed us to both process the data in reasonable time and conduct many iterations to experiment with different model parameters and convergence criteria.

## 2 METHODOLOGY OVERVIEW

### 2.1 Dataset

The DEAP dataset is a large dataset from a controlled setup experiment, where each subject is exposed to a number of pre-rated one-minute extracts of music videos, selected to provoke different emotional states. During the experiment, the electroencephalogram (EEG), peripheral physiological sensors and video, in some cases, are collected, along with the individuals' rankings on valence, arousal, dominance, familiarity and liking. The biosignals for the preprocessed part contain 40 physiological channels (EEG and peripheral signals). The preprocessing includes artifact and baseline removal, filtering, reordering and down-sampling. A detailed description of the dataset can be found in the DEAP-dataset description.

### 2.2 Problem Setup

Figure 1 shows a typical block diagram of the standard classification flow in the training phase, where the labels are used to tune the algorithm's parameters and to build the model, as well as in the prediction phase, where the labels are regenerated for each new data sample. Note that unsupervised learning can be used to augment feature extraction and the output of the unsupervised model can be used as the input to the classification module.

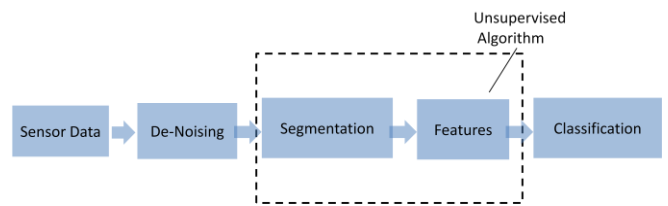

Fig.1. Typical block diagram for a classification problem.

In our case, the data have been preprocessed, so there is no need to use additional denoising methods. In lieu of feature extraction, we used the output of a hard clustering K-means model. The cluster assignments from K-means was then be used as input to the random forests classifier. Our input is the physiological data for 32 subjects on 40 one-minute music videos, from 40 bio-signal channels; leading to an input vector greater than 10 million rows, (8064 x32 subjectsx40 clips), of 40 columns each (one for each physiological sensor), and creating an input file of several Gigabytes. We processed the input file in an unsupervised fashion, using K-means clustering and used its output to augment the classification process.

―――――――――――――――
• *Authors are with Intel Corp. 2200 Mission College Blvd, Santa Clara, CA 95054. E-mail: Varvara.Kollia@intel.com, Oguz.H.Elibol@intel.com*

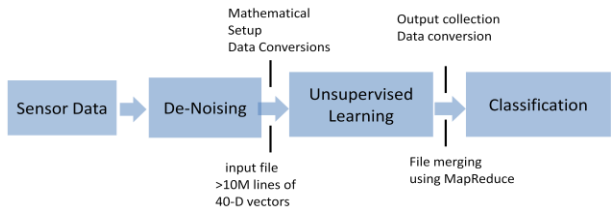

Fig.2. Classification block diagram for distributed processing.

This process is illustrated in Figure 2. A lot of data processing, data conversions and file joining is required, to convert the data into Mahout-compatible, as well as human-readable formats.

The self-assessments of the subjects are used to generate the labels for the classification module. As ground-truth here, we use the self-assessments of the subjects; i.e. the rankings they gave to each video. The rankings, reported on a scale of 1 to 9, for the 3D problem of valence/arousal/dominance are used to generate 8 labels, corresponding to the binary representations for all combinations of whether the corresponding value for each of the three dimensions is greater than the mid point, (4.5), which would lead to a value of (1) or not (0), in the corresponding location of each axis binary representation. The reason behind this mapping is to simplify the problem, while still maintaining the 3D information, in addition to being intuitive and easy to formulate mathematically. A 3D map of emotion in these projections can be seen in Figure 3. In the xy plane, the horizontal axis is valence and the vertical one is arousal. The axis of dominance corresponds to the dotted-lined diagonals.

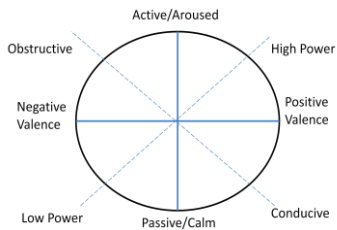

Fig.3. 3D map of emotion.

## 3 RESULTS

### 3.1 K-Means Clustering

In the first stage, the input was pre-processed to merge the independent data files in two files. The first file contained the combined data from all subjects and the second file had all the combined data, as the first file, and, additionally, the corresponding labels. As the clustering algorithms require comparable inputs (for convergence, performance and accuracy reasons), the input data are normalized (with respect to each subject and channel) to have zero mean and unit variance.

Consequently, the input files are vectorized to be compatible with the Mahout input format. The vectors are dense vectors, due to the nature of the data.

The K-means clustering is a hard clustering algorithm, that assigns each vector to one cluster exclusively, based on some similarity criterion. The number of clusters is pre-defined, considered to be one of the algorithm's inputs. Other parameters include the number of iterations, the convergence threshold and the distance (similarity) metric.

We chose the number of clusters to be equal to the number of labels (8) and experimented with different distance measures (Tanimoto, Manhattan, Euclidean, Cosine and Squared Euclidean). The clusters were initialized randomly, based on the input samples. Due to the large number of dimensions, it was hard to visualize the output. More accurate classification results were obtained via the Euclidean distance measure. Regarding the performance of the distributed clustering, 10 iterations on our 5-node cluster required only 25 min, 2 min for each iteration plus 5 min overhead. Each machine has 50 GB disk and 10 GB memory.

The output of the Mahout implementation of the K-means algorithm (centroids and clustered points) was post-processed (bash scripting) to get the human-readable format from the Mahout dense vectors, while storing the centroids and the corresponding data points separately. Furthering processing was required to convert the processed K-means output to a format equivalent to the one of the second input file, so that eventually the two files could be merged and used as input to the random forests classifier. For all the coding parts of this analysis Java7 was used, as Mahout is a Java-based library.

### 3.2 Random Forest Classification

The Random Forest classifier builds its model by training a large number of decision trees and makes decisions taking a majority vote from all the trees. The Out-Of-Bag error (OOB) is used to evaluate the classification error on the training set, in lieu of the prediction error on a test set. The reason is that its training is based on a randomly sampled subset of the data, different for each tree and each seed (initialization) and each sample is classified by taking a majority vote on the trees, for which it was excluded for training. In this manner, the OOB corresponds to the mean prediction error of the confusion matrix.

Before using the output of the K-Means Clustering algorithm, we need to add the label of each data sample to the clustered points file. This is illustrated in Figure 4. Each line corresponds to an input data sample. The first file (rectangle) contains the cluster assignment and the second one the labels, which are represented as integers in the last columns, respectively. The joining operation should match the data of each file and add the label information, in addition to the cluster assignment.

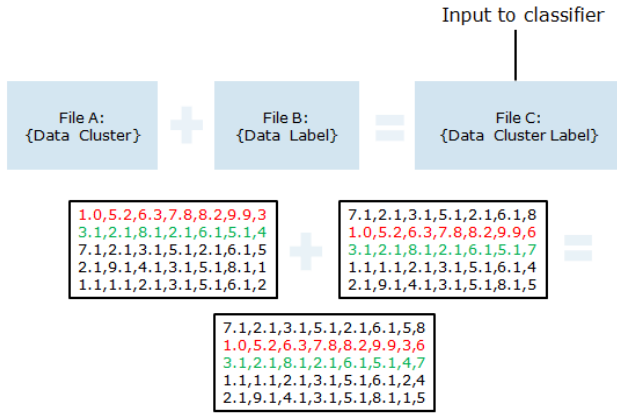

Fig.4. Joining of K-means assignments and labels.

The process of joining these two files, based on their (common) data-field, can take $O(n)$ in the best case, where we find the line we look for with one search only, (i.e. it is the first line every time) up to $O(n^2)$, where we have to look up the entire second file for each line of the first file, (i.e. it is the last line), with $n$ the number of input lines. This is because there are no rules we can take advantage of on how these files were written, so we have to look exhaustively for each line of the first file in all the lines of the second file, until we match each line of the first file. In our case $n$ is larger than $10^7$ and this task takes several days to be completed locally.

Processing these files on Hadoop with MapReduce is finished in less than 8 minutes, as shown in Figure 5. In particular, using the <*key, value*> paradigm, the files are joined based on their *key*, which is the common data field for each input example and the *value* of the output file is the joining of the two values from the individual input files (namely the cluster and the label).

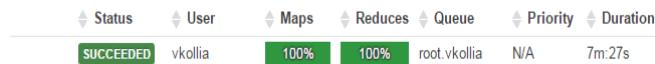

Fig.5. Time for completion for joining two files (>10M inputs).

Finally, in Table I, random forest classification results are presented, in terms of accuracy and reliability and in Table II the accuracies for the individual classes. Given that only unsupervised clustering results were used as input features and the problem was formulated as an 8-label problem, the results demonstrate satisfactory accuracy, compared to literature results. The classification can be augmented with additional input features and a simplified setup. From Table II, we see that the lowest percentage accuracy is observed for the highest values of valence/arousal/dominance. The classes are numbered in increasing order, with respect to their binary representation, namely {0,0,0} is Class1 and {1,1,1} is Class8. The numbering starts from 1. In the binary form, each digit represents whether the corresponding value of {valence, arousal, dominance} is smaller than the midpoint of ranking or not. Regarding the class accuracies, we see that there is large difference among classes. The classes that are more difficult to predict correspond to a smaller number of training samples, in most cases. This is expected, as Random Forest tends to favor the majority classes. Additionally, these classes (i.e. the ones corresponding to all high values for 3D emotion mapping) can be thought of as outliers in the classifying process.

Table I: Random Forests Classifier Results

| Accuracy | 63.3% |
|---|---|
| Reliability | 46.7% |
| Std. Dev. (Reliabil) | 0.33 |

Table II: Individual Class Accuracies

| | Class 1 | Class2 | Class 3 | Class 4 |
|---|---|---|---|---|
| Accuracy(%) | 86.5 | 76.9 | 33.8 | 63.1 |
| | Class5 | Class6 | Class7 | Class8 |
| Accuracy(%) | 75.4 | 44.1 | 73.5 | 14.0 |

## 4 SUMMARY

In this work, we analyze the data from the DEAP dataset for emotion recognition, using EEG and peripheral physiological signals. The data were analyzed in distributed mode with Mahout. All processing was done on Hadoop and the problems were formulated in a way that would fit the MapReduce framework for parallel processing. Unsupervised methods were used to generate the inputs to the random forests classifier, achieving an (OOB-based) accuracy of ~63% on the training set for 8-label classification. The performance in terms of time is not even comparable to the local setting, as operations that take days to execute locally conclude in minutes in a small cluster.

## 5 REFERENCES

[1] P. Ekman, *Facial expression and emotion*, American Psychologist, vol. 48, no. 4, 1993, pp. 384–392.
[2] J. A. Russell, *A circumplex model of affect*, Journal of Personality and Social Psychology, vol. 39, no. 6, 1980, pp. 1161–1178.

[3] K. R. Scherer, *What are emotions? And how can they be measured*, Social Science Information, vol. 44, no. 4, 2005, pp. 695-729.

[4] S. Koelstra, C. Muehl, M. Soleymani, J.-S. Lee, A. Yazdani, T. Ebrahimi, T. Pun, A. Nijholt, I. Patras, *DEAP: A Database for Emotion Analysis using Physiological Signals*, IEEE Trans. on Affective Computing, vol. 3, no 1, 2012, pp.18-31.

[5]DEAP/Dataset:
http://www.eecs.qmul.ac.uk/mmv/datasets/deap/ .

[6] J. G. Proakis and D. G. Manolakis, Digital Sinal Processing, 4th Ed., Pearson, 2006.

[7] L. Breiman, *Random Forests*, Article Machine Learning, Kluwer Academic Publishers, vol 45, issue 1, 2001, pp. 5-32.

[8] T. Hastie, J. H. Friedman, R. Tibshirani, The Elements of Statistical Learning, 2nd Ed, Springer, 2001.

[9] http://www.cloudera.com/ .

[10] http://stackoverflow.com/.

[11] http://mahout.apache.org/ .

[12]http://www.eclipse.org/downloads/packages/release/Luna/SR2 .

[13] V. Kollia, *Personalization effect on Emotion Recognition: An Investigation of Performance on Different Setups and Classifiers*, https://arxiv.org/abs/1607.05832 .

[14] http://www.oracle.com/technetwork/java/ .

[15] https://linuxconfig.org/bash-scripting-tutorial .

[16] S. Owen, R. Anil, T. Dunning, E. Friedman, Mahout in Action, Manning Publications Co., 2012, https://github.com/tdunning/MiA .

[17] P. Srinath, Instant MapReduce Patterns-Hadoop essentials How-to, PACKT Publishing, 1st ed., 2013.

[18] D. Wang and Y. Shang, *Modeling Physiological Data with Deep Belief Networks*, Int J Inf Educ Tehnol. 2013, 3(5), pp. 505-511, doi:10.7763/IJIET.2013.V3.326.

[19] X. Li, P. Zhang, D. Song, G. Yu, Y. Hou, B. Hu, "*EEG Based Emotion Identification Using Unsupervised Deep Feature Learning*", in SIGIR2015 Workshop on NeuroPhysiological Methods in IR Research, Santiago, Chile, 2015.